\pdfoutput=1

\documentclass{article}
\usepackage{ijcai23}

\usepackage{times}
\usepackage{soul}
\usepackage{url}
\usepackage[hidelinks]{hyperref}
\usepackage[utf8]{inputenc}
\usepackage[small]{caption}
\usepackage{graphicx}
\usepackage{amsmath}
\usepackage{amsthm}
\usepackage{booktabs}
\usepackage{algorithm}
\usepackage{algorithmic}
\usepackage[switch]{lineno}
\usepackage{amssymb}
\usepackage{booktabs}
\usepackage{float}
\usepackage{multirow}
\usepackage{subfigure}
\usepackage{newfloat}
\usepackage{listings}
\usepackage{url}
\usepackage[table]{xcolor}
\usepackage{color,xcolor}
\definecolor{darkgreen}{rgb}{0,0.5,0} 

\usepackage[title]{appendix}

\hypersetup{
    colorlinks=true,
    urlcolor=magenta,
}

\usepackage{xcolor}         

\title{Detecting Adversarial Faces Using Only Real Face Self-Perturbations}

\author{
Qian Wang$^1$\and
Yongqin Xian$^2$\and
Hefei Ling$^{1,}$\thanks{Corresponding author}\and
Jinyuan Zhang$^3$\and\\
Xiaorui Lin$^3$\and
Ping Li$^1$\and
Jiazhong Chen$^1$\and
Ning Yu$^4$
\affiliations
$^1$Huazhong University of Science and Technology, Wuhan, China \\
$^2$Google, Switzerland\\
$^3$Software Development Center, Industrial and Commercial Bank of China\\ 
$^4$Salesforce Research, USA
\emails
$^1$\{yqwq1996, lhefei, lpshome, jzchen\}@hust.edu.cn, \quad
$^2$yxian@google.com,\\
$^3$\{zhangjy, linxr\}@sdc.icbc.com.cn, \quad
$^4$ning.yu@salesforce.com
}

\begin{document}

\maketitle

\begin{abstract}
Adversarial attacks aim to disturb the functionality of a target system by adding specific noise to the input samples, bringing potential threats to security and robustness when applied to facial recognition systems.
Although existing defense techniques achieve high accuracy in detecting some specific adversarial faces (adv-faces), new attack methods especially GAN-based attacks with completely different noise patterns circumvent them and reach a higher attack success rate. 
Even worse, existing techniques require attack data before implementing the defense, making it impractical to defend newly emerging attacks that are unseen to defenders. 
In this paper, we investigate the intrinsic generality of adv-faces and propose to generate pseudo adv-faces by perturbing real faces with three heuristically designed noise patterns. 
We are the first to train an adv-face detector using only real faces and their self-perturbations, agnostic to victim facial recognition systems, and agnostic to unseen attacks. 
By regarding adv-faces as out-of-distribution data, we then naturally introduce a novel cascaded system for adv-face detection, which consists of training data self-perturbations, decision boundary regularization, and a max-pooling-based binary classifier focusing on abnormal local color aberrations. 
Experiments conducted on LFW and CelebA-HQ datasets with eight gradient-based and two GAN-based attacks validate that our method generalizes to a variety of unseen adversarial attacks. 
\footnote{Code at \href{https://github.com/cc13qq/SAPD}{https://github.com/cc13qq/SAPD}}
\end{abstract}

\section{Introduction}

\begin{figure}[ht]
    \centering
    \includegraphics[width=0.9\columnwidth]{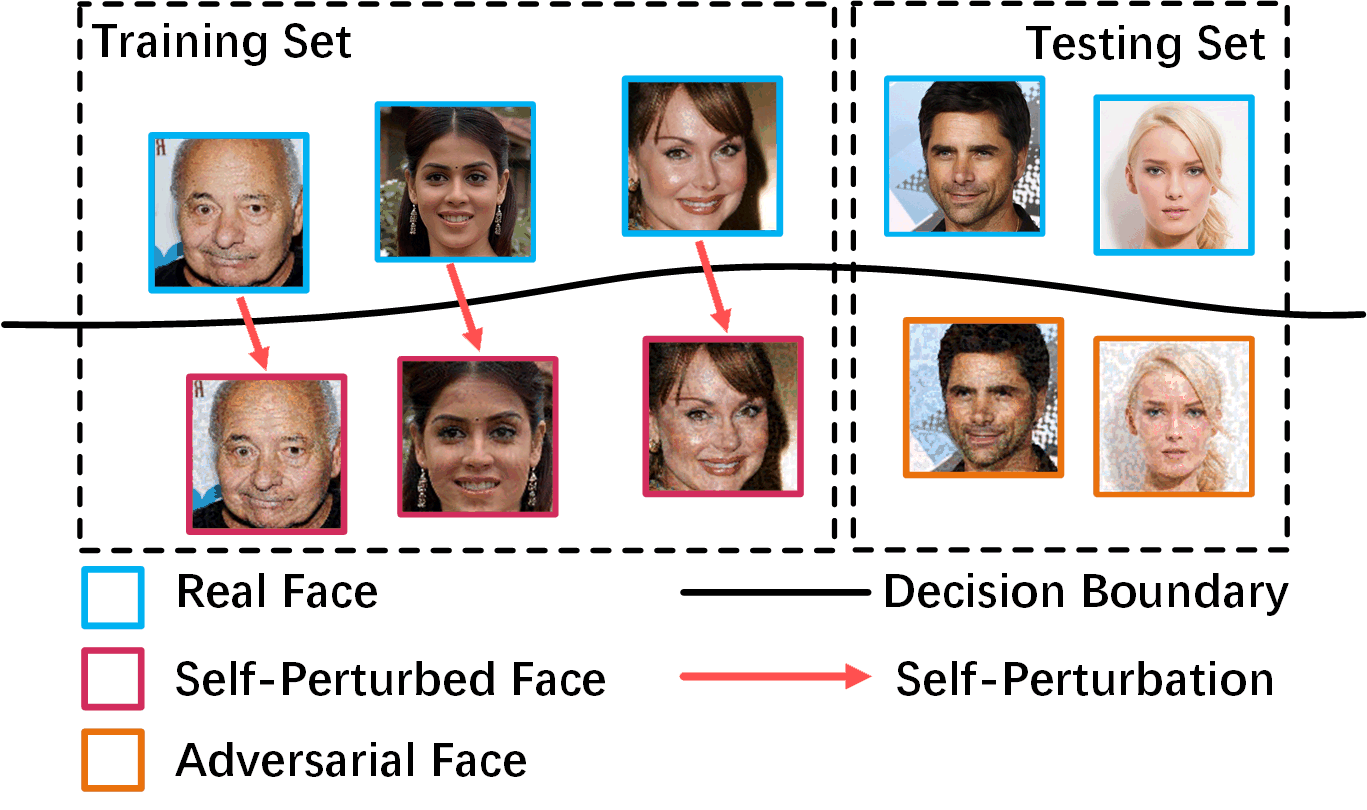}
    \caption{
    Trained only on real faces and their self-perturbed faces, the detector learns a generic representation of adversarial faces produced by unseen attacks.}
    \label{fig: teaser_figure}
\end{figure}

Deep neural networks have been widely used in many tasks \cite{Ou:17Adult,Shi:20Learning,Zhang:21Mixture}, and have achieved remarkable success in facial recognition systems (FRS)~\cite{Arcface} with a wide range of real-world applications such as online payment \cite{OnlinePayment} and financial management \cite{FinancialManagement}. 
However, deep neural networks are known to be vulnerable to adversarial attacks~\cite{FGSM,BIM,PGD}, making commercial FRSs unreliable.

\begin{figure*}[t]
    \centering
    \includegraphics[width=2.1\columnwidth]{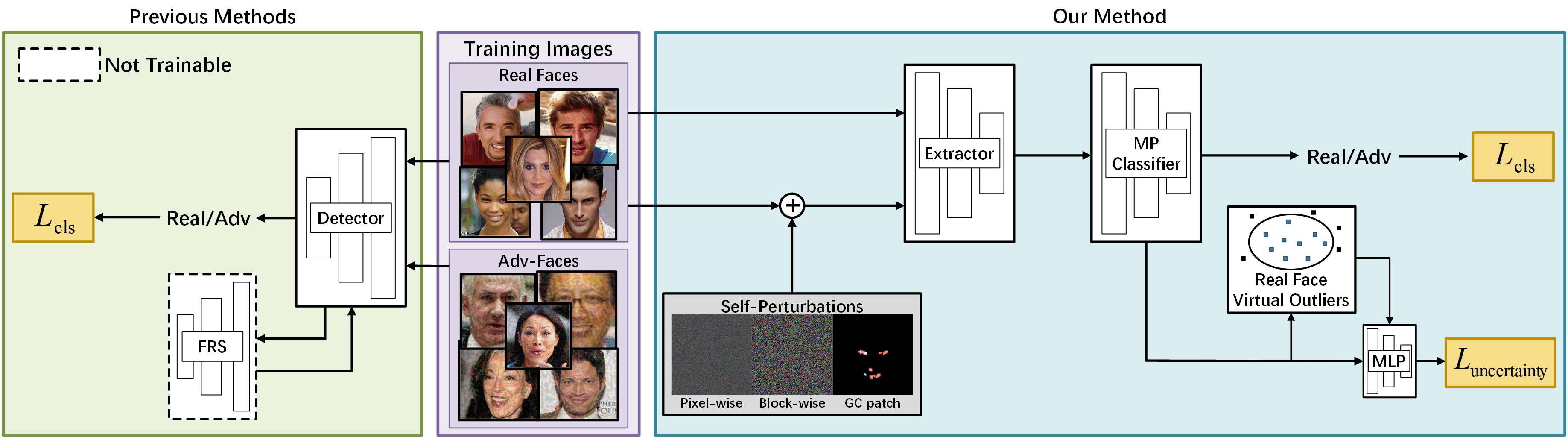}
    \caption{Overview of training pipeline. We design three kinds of self-perturbations to generate synthesized pseudo-adv faces. "GC patch" denotes a gradient color patch for GAN-based attack. "MP Classifier" denotes Max-Pooling Classifier which focuses on abnormal local color aberrations. By synthesizing virtual outliers of real faces, we incorporate an uncertainty loss to regularize the decision boundary to enhance detection performance on adv-faces. Previous methods require pre-computed adv-faces and access permission to the victim FRS for training. By contrast, our method is agnostic to both attacks and FRS. Using only real faces with self-perturbations in the training phase, our method is able to detect all kinds of adv-faces from unseen attacks without modifying or visiting FRS.}
    \label{fig: training_pipeline}
\end{figure*}

Some research attempts to defend against adversarial attacks by adversarial example detection techniques \cite{KD,LID,SID} which filter out adversarial samples before feeding them into protected systems. 
However, they tend to overfit to known attacks and do not generalize well to unseen advanced attacks \cite{DIFGSM,TIPIM}. 
In specific, GAN-based adversarial attacks such as AdvMakeup \cite{AdvMakeup} and AMT-GAN \cite{AMT-GAN} are recently developed to generate more natural adversarial faces (adv-faces) by adding completely different noise patterns to face images in contrast to traditional gradient-based attacks such as FGSM \cite{FGSM}, and are able to circumvent detection. 
Therefore, a plug-and-play method of detecting adv-faces with improved generalization performance over both unseen gradient-based attacks and unseen GAN-based attacks is highly demanded.

Due to the insight that adversarial samples are all modified from real samples \cite{LID}, one direct way to encourage models to learn generic representations for detecting adv-faces is to train models with synthetic data \cite{SBIs}, forming a decision boundary wrapping the synthetic subspace. 
To achieve this, we investigate the intrinsic generality of adv-faces generated by gradient-based attacks and GAN-based attacks respectively, and propose two assumptions: 
(1) Gradient-based attacks have a resemblance in noise pattern on account of the consistency of the basic attack algorithm which uses gradients to construct adversarial examples.
(2) GAN-based attacks aim to simultaneously change the prediction of FRSs while preserving visual quality. In the meantime, FRSs pay the most attention to high-frequency regions \cite{FaceRecognition},  
which induces GANs to modify these regions. Therefore, GAN-made adv-faces have manipulated clues like abnormal color aberrations in high-frequency regions. 
If universal noise patterns that cover all kinds of adversarial noises are attainable, a set of pseudo adv-faces could be produced, and we can train a model with them to learn generic representations for adv-faces. 

Based on the above assumptions, we propose three kinds of real face self-perturbations synthesized with pseudo noise patterns which summarize adversarial noises of gradient-based attacks and GAN-based attacks, and a max-pooling-based classifier focusing on capturing abnormal local color aberration. 
Unlike previous methods, this data-augmentation-like method is proposed from a novel and more intrinsic perspective by investigating the generality of adversarial noise patterns, to detect all recent adv-faces from unseen attacks, protecting FRSs without access to them.
Trained on only real faces and their self-perturbations, as shown in Figure \ref{fig: teaser_figure}, our model is able to detect all the recent adv-faces. This new framework simultaneously ensures detection performance on unseen adversarial attacks and portability in use, without any access to protected systems. 


Although we are trying to make self-perturbed faces as general as possible, the trained network still overfits to in-domain data and might fail to classify adv-faces far away from the distribution of self-perturbed faces. 
Regarding adv-faces as out-of-distribution (OOD) data from a shifted data manifold \cite{LiBRe}, we incorporate a regularization term to narrow
the model’s decision boundary of real face class during training, and naturally introduce a novel cascaded system for adv-face detection, which consists of 
training data self-perturbations, decision boundary regularization, and max-pooling-based binary classifier.
We evaluate our approach on LFW and CelebA-HQ datasets with eight gradient-based attack methods and two GAN-based attack models. Experiment results validate our assumptions and demonstrate the reliability of our detection performance on adv-faces from unseen adversarial attack methods. 

Our contributions are summarized in four thrusts:

$\bullet$ By investigating the intrinsic similarities among varying adv-faces, we propose two assumptions that (1) different adversarial noises have resemblances, and (2) color aberrations exist in high-frequency regions. We empirically validate the assumptions which in turn indicate universal noise patterns are attainable for all the recent adv-faces. 

$\bullet$ Based on our assumptions, we propose three kinds of real face self-perturbations for gradient-based adv-faces and GAN-based adv-faces. We are the first to train an adv-face detector using only real faces and their self-perturbations, agnostic to victim FRSs and agnostic to adv-faces. This enables us to learn generalizable representations of adv-faces without overfitting to any specific one.

$\bullet$ We then naturally introduce a novel cascaded system for adv-face detection, which consists of 
training data self-perturbations, decision boundary regularization, and a max-pooling-based binary classifier focusing on abnormal local color aberrations.

$\bullet$ Evaluations are conducted on LFW and CelebA-HQ datasets with eight gradient-based attacks and two GAN-based attacks, demonstrating the consistently improved performance of our system on adv-face detection.

\section{Related Work}

\paragraph{Adversarial attacks}
Adversarial attacks \cite{FGSM} aim to disturb a target system by adding subtle noise to input samples, while maintaining imperceptibility from human eyes. 
In contrast to previous works \cite{PGD,BIM}, DIM \cite{DIFGSM} and TIM \cite{TIFGSM} are improved attack algorithms with enhanced black-box attack accuracy and breaching several defense techniques. 
Focusing on breaking through FRSs, TIPIM \cite{TIPIM} generates adversarial masks for faces to conduct targeting attacks while remaining visually identical to the original version for human beings. 
Recently, GAN-based attack models \cite{AdvMakeup,AMT-GAN} are presented to generate adversarial patches and imperceptible adv-faces, bringing new challenges to adv-faces detection. 
In this paper, we proposed a simple yet effective detection method against all the above attacks while being blind to them during training. 


\paragraph{Adversarial example detection}
One of the technical solutions for protecting DNNs from adversarial attacks is adversarial example detection \cite{KD}, aiming to filter out adversarial inputs before the protected system functions. 
Remedying the limitations of the previous method, LID \cite{LID} is proposed for evaluating the proximity of an input to the manifold of normal examples. 
Incorporated with wavelet transform, SID \cite{SID} is able to transform the decision boundary and can be collaboratively used with other classifiers. 
While all of these methods focus on utilizing features subtracted by FRS backbone and show effective performance in detecting adversarial examples, they require modifying or visiting the protected system and do not generalize well to new-type attacks like GAN-based AMT-GAN \cite{AMT-GAN} and AdvMakeup \cite{AdvMakeup}.
To address the problems above, our method is proposed to detect all recent adv-faces from unseen attacks, protecting FRSs without access to them. 

\paragraph{Out-of-distribution (OOD) detection}
OOD detection techniques \cite{OpenGAN} have been widely used in image classification tasks \cite{GANfingerprint}, trying to recognize examples from an unknown distribution. 
ODIN \cite{ODIN} uses temperature scaling and tiny perturbations to the inputs to separate the in-distribution (ID) and OOD images. 
Lee et al. \cite{MD} uses the Mahalanobis distance to evaluate the dissimilarity between ID and OOD samples.
By sampling and synthesizing virtual outliers from the low-likelihood regions, VOS \cite{VOS} adaptively regularizes the decision boundary during training. 
Sun et al. \cite{ReAct} introduce a simple and effective post hoc OOD detection approach utilizing activation truncation.
Regarding adv-faces as OOD data, we leverage an uncertainty regularization term to narrow the decision boundary of real face class during the training phase to boost the accuracy of adv-faces detection.

\section{Methodology}

\begin{algorithm}[ht]
\caption{Self-perturbation for gradient-based attack}
\label{alg:pseudo gradient}
\textbf{Input}: A empty perturbation matrix $\boldsymbol{\eta}^p \in \mathbb{R}^{H\times W\times3}$ with the same shape of real face image $\mathbf{x}^r$.\\
\textbf{Parameter}: Max perturbation magnitude $\epsilon$, pattern mode.\\
\textbf{Output}: Self-perturbed face image $\mathbf{x}^p$.

\begin{algorithmic}[1] 
\STATE A random direction matrix $R = \{\vec{r}_{ij} \} \in \mathbb{R}^{H\times W\times3}$.
\FOR{$\vec{\eta}_{ij}$ \textbf{in} $\boldsymbol{\eta}^p$}
\STATE Select random noise value $\alpha$.
\IF{pattern mode \textbf{is} ‘point-wise’}
\STATE $\vec{\eta}_{ij} := \alpha \cdot \vec{r}_{ij}$.
\ELSIF{pattern mode \textbf{is} ‘block-wise’}
\STATE Select a random neighborhood $A_{ij}$ of $\vec{\eta}_{ij}$.
\FOR{$\vec{\eta}_{ijk}$ \textbf{in} $A_{ij}$}
\STATE $\vec{\eta}_{ijk} := \alpha \cdot \vec{r}_{ij}$.
\ENDFOR
\ENDIF
\ENDFOR
\STATE Clip perturbation $\boldsymbol{\eta}^p$ using Equation \ref{eq:noise clip}.
\STATE Generate self-perturbed face $\mathbf{x}^p$ using Equation \ref{eq:pesudo AE}.
\STATE \textbf{return} $\mathbf{x}^p$
\end{algorithmic}

\end{algorithm}

We propose a plug-and-play cascaded system for adv-faces detection method which consists of 
training data self-perturbations, decision boundary regularization, and a max-pooling-based binary classifier focusing on abnormal local color aberrations, agnostic to unseen adversarial attacks, and agnostic to victim FRSs. 
Training pipeline as shown in Figure \ref{fig: training_pipeline}.
We propose to synthesize diverse self-perturbed faces by adding three noise patterns to real face images, summarizing adv-faces generated from gradient-based and GAN-based attacks. 
A convolutional neural network is learned via real faces and self-perturbed faces, regularized by uncertainty loss, to distinguish real and adversarial faces.
As no attack method is observed during training, the resulting network is not biased to any attack, yielding generic and discriminative representations for detecting unseen adv-faces. 
Therefore, in the testing phase, the embeddings are sent to the learned Max-Pooling Classifier to accomplish prediction. 


\subsection{Real Face Self-Perturbations}
\paragraph{Self-perturbation for gradient-based attack}
Given a real face image $\mathbf{x}^r$, 
adversarial attack generates an adversarial image $\mathbf{x}^a$ by adding a perturbation image $\boldsymbol{\eta}$ to $\mathbf{x}^r$ \cite{FGSM}:
\begin{equation}
    \mathbf{x}^a = \mathbf{x}^r + \boldsymbol{\eta},
\end{equation}
where $||\boldsymbol{\eta}|| \leq \epsilon$, and $\epsilon$ called perturbation magnitude.

We observe that a binary classifier trained on real faces and adv-faces generated by FGSM is able to classify a part of attack images generated from other attack methods as shown in Table \ref{tab: attack cross}. 
This generalization in classification means that attack noises have intrinsic similarities. And it means once we master all the similarities we master the attacks, even for unseen attacks. This leads to non-trivial designs for our self-perturbation image $\boldsymbol{\eta}^p$. 

The noise value of neighbor points in $\boldsymbol{\eta}^p$ may be the same or different. From this perspective, we introduce point-wise and block-wise noise patterns for gradient-based attacks. 
As presented in Algorithm \ref{alg:pseudo gradient}, 
we perturb each point in the point-wise pattern and each block in the block-wise pattern in a stochastic direction,
where blocks are random neighborhoods of a set of scattered points. The generated perturbation image $\boldsymbol{\eta}^p$ is constrained in $l^\infty$ norm, and clipped according to $\epsilon$,
\begin{equation}
\label{eq:noise clip}
    \boldsymbol{\eta}^p = \mathrm{Clip}_{[-\epsilon, \epsilon]}(\boldsymbol{\eta}^p).
\end{equation}
And self-perturbed face is calculated as
\begin{equation}
\label{eq:pesudo AE}
    \mathbf{x}^p = \mathbf{x}^r + \boldsymbol{\eta}^p.
\end{equation}

\begin{algorithm}[ht]
\caption{Self-perturbation for GAN-based attack}
\label{alg:pseudo GAN}
\textbf{Input}: Real face image $\mathbf{x}^r$. A empty perturbation matrix $\boldsymbol{\eta}^p = \{\eta_{ij}\}$ with the same shape of $\mathbf{x}^r$.\\
\textbf{Parameter}: Max perturbation magnitude $\epsilon$, face landmarks, high-frequency threshold $\gamma$, and empty high-frequency pixel set $H$.\\
\textbf{Output}: Self-perturbed face image $\mathbf{x}^p$.\\
\textbf{Function}: $\mathrm{value}(x)$ measures pixel value of $x$, $\mathrm{neighborhood}(x)$ is a random neighborhood of $x$.

\begin{algorithmic}[1] 
\STATE Obtain gradient image of $\mathbf{x}^r$ using Sobel operator.
\STATE Obtain high-frequency convex hull from gradient image according to landmarks.
\FOR{$x_{ij}$ \textbf{in} convex hull}
\IF{$\mathrm{value}(x_{ij}) < \gamma$}
\STATE $\eta_{ij}$ join to $H$.
\ENDIF
\ENDFOR
\STATE Random select a subset $H_s$ of $H$.
\FOR{$\eta^H_{ij}$ \textbf{in} $H_s$}
\STATE Generate a random gradient color patch $P$.
\STATE $\mathrm{neighborhood}(\eta^H_{ij}) := P$.
\ENDFOR
\STATE Clip perturbation $\boldsymbol{\eta}^p$ using Equation \ref{eq:noise clip}.
\STATE Generate self-perturbed face $\mathbf{x}^p$ using Equation \ref{eq:pesudo AE}.
\STATE \textbf{return} $\mathbf{x}^p$
\end{algorithmic}
\end{algorithm}

\paragraph{Self-perturbation for GAN-based attack}
Spectrums of real and fake images distribute differently in high frequency \cite{YuchenLuo2021}, such as eyes, nose, and mouth. 
This distribution difference should become more intense because GAN-made adv-faces aim to change the classification results of FRSs meanwhile maintain perceptual invariance. 
This leads to color aberration and boundary abnormality in high-frequency regions on account of that FRSs pay the most attention to these regions. 
Focusing on producing natural and imperceptible attack noise, these abnormal color aberrations should have gradient or blurred boundaries. 
In such a perspective, we use gradient color patches to act as self-perturbation.

As presented in Algorithm \ref{alg:pseudo GAN}, we first obtain the gradient image of a real image using Sobel operator \cite{SobelOperator} and convex hull of the high-frequency area according to facial landmarks. 
The pixels in the convex hull are selected as high-frequency pixels if their values surpass a threshold $\gamma$. 
Then we generate a few gradient color patches through a series of affine transformations.
Finally, the pseudo adv-faces are generated by adding gradient color patches to a random set of high-frequency pixels. 

After crafting a part of real faces to self-perturbed faces as negative samples, we label other real faces as positive samples and train a well-designed backbone network such as XceptionNet \cite{XceptionNet} in a binary classification manner.

\subsection{Max-Pooling Classifier}
In view of abnormal speckles in GAN-based adv-faces are always tiny and are not easy to be observed, we proposed Max-Pooling Classifier (MPC) to capture abnormal local color aberrations. MPC produces classification scores and predicts whether input images are real or adversarial. 

A typical backbone network includes a feature extractor $f(\cdot ;\boldsymbol{\varphi})$ composed of several convolutional blocks, and a classifier $g(\cdot ;\boldsymbol{\phi})$ comprising an average pooling layer, an activation layer, and a fully connected layer \cite{XceptionNet}. 
To detect adv-faces with abnormal local color aberrations, a simple derivation is to divide an image into several rectangle areas, and if abnormal color aberration occurs in any of the rectangle areas we classify this image as adv-face. 

Input an image $\mathbf{x}$ to feature extractor $f(\cdot ;\boldsymbol{\varphi})$, it produces a feature map $M(\mathbf{x})\in \mathbb{R}^{N\times N\times d}$ composed of $N\times N$ features, where $d$ is the embedding size, and each feature is corresponding to a rectangle area i.e. the receptive field. 
Features are sent to the activation function and fully connected layer respectively and produce $N\times N$ logits. We take the max-pooling of these logits as the final logit and obtain prediction score $S_{\text{cls}}\in \mathbb{R}^{2}$ by softmax operation. 
Classification of input image $\mathbf{x}$ is computed by argmax function: 
\begin{equation}
     G_{\text{cls}}(\mathbf{x}) = \mathrm{argmax}(S_{\text{cls}}),
\end{equation}
where $G_{\text{cls}}(\mathbf{x}) = 1$ indicates real face and $G_{\text{cls}}(\mathbf{x}) = 0$ indicates adv-face. 


\subsection{Decision Boundary Regularization}

Although we are trying to make self-perturbed faces as general as possible, the trained network still overfits to ID data and might fail to classify faces far away from the distribution of negative samples.  
Regarding adv-faces as OOD data, we incorporate a regularization term \cite{VOS} to our work to enhance adv-face detection performance. 

Assuming the feature representation of real faces forms a multivariate Gaussian distribution, we sample virtual outliers $\mathcal{V} \subset \mathbb{R}^m$ from the $\varepsilon$-likelihood region of the estimated class-conditional distribution:
\begin{equation}
    \mathcal{V}=\left\{\mathbf{v} \mid \frac{\textrm{exp}\left(-\frac{1}{2}\left(\mathbf{v}-\widehat{\boldsymbol{\mu}}\right)^{\top} \widehat{\boldsymbol{\Sigma}}^{-1}\left(\mathbf{v}-\widehat{\boldsymbol{\mu}}\right)\right)}{(2 \pi)^{m / 2}|\widehat{\boldsymbol{\Sigma}}|^{1 / 2}} <\varepsilon\right\},
\end{equation}
where $\widehat{\boldsymbol{\mu}}$ and $\widehat{\boldsymbol{\Sigma}}$ are the estimated mean and covariance using latent features of real faces.

The uncertainty loss regularizes the model to produce a low OOD score for ID data and a high OOD score for the synthesized outliers, and narrows the decision boundary of real face class to boost the performance of adv-face detection:
\begin{equation}
\begin{aligned}
    \mathcal{L}_{\text{uncertainty}}&=\mathbb{E}_{\mathbf{v} \sim \mathcal{V}}\left[\log (\exp ^{-\phi(- f(\mathbf{v}; \theta))}+1)\right]\\
    &+\mathbb{E}_{\mathbf{x} \sim \mathcal{D}^r}\left[\log (\frac{1}{\exp ^{-\phi(- f(\mathbf{x}; \theta))}}+1)\right] ,
\end{aligned}
\end{equation}
where $\mathcal{D}^r$ represents distribution of real face, $f(\mathbf{\cdot}; \theta)$ is a linear transformation function, and $\phi(\cdot)$ is a nonlinear MLP function. The learning process shapes the uncertainty surface, which predicts a high probability for ID data and a low probability for virtual outliers $\mathbf{v}$.

\begin{table*}[ht]
    \centering
    \tabcolsep=0.05cm
    \begin{tabular}{c|lllccc}
        \toprule
        \multicolumn{2}{c}{\textbf{Study}} & \textbf{Method} & \textbf{Detect Attacks} & \textbf{Attack-Agnostic} & \textbf{FRS-Agnostic} \\
        \midrule
        \multirow{2}{*}{\rotatebox{90}{AD}} & LID \cite{LID} & Local intrinsic dimensionality & Gradient-based & $\times$ & $\times$  \\ 
        \multirow{2}{*}{ } & SID \cite{SID} & Wavelet transform & Gradient-based & $\times$ & $\times$ \\ 
        \hline
        \multirow{2}{*}{\rotatebox{90}{FD}} & Luo et al. \cite{YuchenLuo2021} & SRM convolution + DCMA & GAN-based & $\times$ & \checkmark \\ 
        \multirow{2}{*}{ } & He et al. \cite{YangHe2021} & Re-Synthesis Residuals & GAN-based & $\times$ & \checkmark \\
        \hline
        \multirow{3}{*}{\rotatebox{90}{OOD}} & ODIN \cite{ODIN} & Softmax score & Gradient-based + GAN-based & \checkmark & $\times$  \\
        \multirow{3}{*}{ } & MD \cite{MD} & Mahalanobis distance. & Gradient-based + GAN-based & \checkmark & $\times$ \\
        \multirow{3}{*}{ } & ReAct \cite{ReAct} & Rectified truncation & Gradient-based + GAN-based & \checkmark & $\times$  \\
        \midrule
        \multicolumn{2}{c}{\textbf{Ours}} & Self-Perturbation & Gradient-based + GAN-based & \checkmark & \checkmark  \\
        \bottomrule
    \end{tabular}
    \caption{Baselines used in our study. "AD" and "FD" denotes adversarial example detection and face forgery detection respectively. "OOD" denotes OOD detection. "Attack-Agnostic" means that attacks are unseen and no adv-faces are required for training. "FRS-Agnostic" means that victim FRS are unknown and access is forbidden. Only our method does not rely on pre-computed adv-faces or victim FRS.} 
    \label{tab: baseline}
\end{table*}

The training objective combines the real-adv classification loss and the regularization term:
\begin{equation}
    \mathrm{min} \mathbb{E}_{(\mathbf{x},y)\sim \mathcal{D}} (\mathcal{L}_{\text{cls}} + \beta \cdot \mathcal{L}_{\text{uncertainty}}),
\end{equation}
where $\mathcal{D}$ represents the distribution of training data. $\beta$ is the weight of the regularization $\mathcal{L}_{\text{uncertainty}}$, and $\mathcal{L}_{\text{cls}}$ is the Cross-Entropy classification loss \cite{CrossEntropyLoss}. 


\section{Experiment}
To validate the detection performance of our approach on adv-faces generated by various adversarial attack methods, we conduct extensive empirical studies on two datasets in this section. We validate our assumptions by investigating the intrinsic similarity of adv-faces and compare our method to baselines from three research streams. After that, we analyze our approach through a series of ablation studies. 

\paragraph{General setup}
During model training, real faces are used as positive samples, and only self-perturbations of real faces are used as negative samples. Any adv-faces are agnostic. During testing, all the negative samples are adv-faces. 
In detail, half of the real faces in the training phase are labeled $1$ as the positive samples, and others are self-perturbed and labeled $0$ as the negative samples.
In the testing phase, all of adv-faces are labeled $0$. 
For evaluating the performance of detectors, we choose the widely used AUC score as the main metric.

\paragraph{Datasets}
Face images in this work are sampled from LFW \cite{LFW} and CelebA-HQ \cite{CelebA-HQ} datasets. 
LFW contains 13,233 face images of 5,749 subjects.
Subjects with at least two face images take part in adv-face producing. 
The first image of each subject is regarded as a reference and the others are sampled to produce adv-faces which includes 7,484 images. 
CelebA-HQ is a high-resolution subset of CelebA, containing 30,000 images.

\paragraph{Attack methods}
We employ eight recent gradient-based adversarial attack methods FGSM \cite{FGSM}, BIM \cite{BIM}, PGD \cite{PGD}, RFGSM \cite{RFGSM}, MIM \cite{MIFGSM}, DIM \cite{DIFGSM}, TIM \cite{TIFGSM} and TIPIM \cite{TIPIM}, and two GAN-based attack methods: AdvMakeup \cite{AdvMakeup} and AMT-GAN \cite{AMT-GAN} as attackers, 
and ArcFace \cite{Arcface} as victim FRS to produce adv-faces on gradient-based attacks. 
All the gradient-based attacks are applied on LFW and CelebA-HQ datasets to generate adv-faces, while AdvMakeup and AMT-GAN are on the CelebA-HQ dataset. 

\begin{table}[t]
    \centering
    \begin{tabular}{l|ccccccc}
        \toprule
        Detector & FGSM & PGD & DIM & TIM & AdvM.\\
        \hline
        FGSM & 100 & 99.9 & 99.7 & 49.2 & 0.0\\
        PGD & 99.4 & 100 & 99.7 & 6.8 & 0.0 \\
        DIM & 98.7 & 99.4 & 99.6 & 2.2 & 0.0 \\
        TIM & 71.3 & 85.8 & 79.5 & 93.2 & 0.0 \\
        AdvM. & 22.1 & 19.9 & 19.5 & 23.2 & 98.5 \\
        \bottomrule
    \end{tabular}
    \caption{Detection accuracy (\%) on adv-faces generated by various of attack algorithms. All adv-faces are generated at $\epsilon = 5/255$. Networks are trained on adv-faces generated by attacks in the left column, and tested on adv-faces produced by attacks in the top row. }
    \label{tab: attack cross}
\end{table}

\paragraph{Implementation details}
We modify an ImageNet-pre-trained XceptionNet \cite{ImageNet,XceptionNet} as the backbone network in our method. 
We set $N=7$ to produce a $7\times 7$ feature map in the last convolution layer and choose ReLU as an activation function. 
We utilize DLIB \cite{DLIB} for face extraction and alignment, Torchattacks \cite{torchattacks} for generating adv-faces, and OpenOOD \cite{OpenGAN} for network training. 
All face images are aligned and resized to $256\times 256$ before training and testing. 
The perturbation magnitude $\epsilon$ in self-perturbations and adv-faces producing is set to $5/255$, a small value. 
Threshold $\gamma$ in the convex hull of gradient image in Algorithm \ref{alg:pseudo GAN} is set to $50$.
The regularization loss weight $\beta$ is set to $0.1$. 
Training epochs are set to 5 and convergence is witnessed.

\paragraph{Baselines}
Previous adversarial example detection methods \cite{LID,SID} focus on specific tasks or gradient-based attacks and hence can hardly be effectively extended to GAN-based adv-faces, while face forgery detection methods \cite{YuchenLuo2021,YangHe2021} are used to detect GAN-made fake faces. 
On the other hand, some OOD detection methods \cite{ODIN,MD,ReAct} only rely on output features and logits of the backbone network and keep agnosticism to unknown attacks, closing to our setting.
On account of this, we compare our method to various methods in three problem settings. 
All baselines listed in Table \ref{tab: baseline} include methods of adversarial example detection, OOD detection, and face forgery detection. 
It is worth noting that only our method does not require either pre-computed adv-faces or access to FRS.

\begin{table*}[ht]
    \centering
    \tabcolsep=0.16cm
    \begin{tabular}{l|cccccccc|cccccccc}
        \toprule
        \multirow{2}*{Method} & \multicolumn{8}{c|}{LFW} & \multicolumn{8}{c}{CelebA-HQ} \\
        \cline{2-17}
        ~ & FGSM & BIM & PGD & RF. & MIM & DIM & TIM & TIP. & FGSM & BIM & PGD & RF. & MIM & DIM & TIM & TIP.\\
        \hline
        LID & 76.7 & 74.0 & 70.7 & 73.0 & 77.7 & 70.2 & 62.1 & 69.0 & 82.0 & 55.2 & 52.5 & 54.4 & 57.7 & 53.7 & 54.0 & 59.3 \\ 
        SID & 99.7 & 81.8 & 73.7 & 77.8 & 90.1 & 72.2 & 73.4 & 88.5 & 96.7 & 79.2 & 63.4 & 72.8 & 84.5 & 76.4 & 81.0 & 85.2 \\ 
        \hline
        ODIN & 75.6 & 71.1 & 71.6 & 75.2 & 79.8 & 73.6 & 74.8 & 71.4 & 76.7 & 75.7 & 75.8 & 75.7 & 75.0 & 76.0 & 72.4 & 72.2\\
        MD & 95.2 & 91.3 & 91.9 & 91.4 & 90.1 & 94.0 & 88.9 & 86.6 & 94.1 & 91.6 & 91.6 & 91.7 & 90.7 & 93.2 & 91.9 & 87.4\\
        ReAct & 92.3 & 89.2 & 88.4 & 89.1 & 89.9 & 92.2 & 91.3 & 90.8 & 93.6 & 90.7 & 90.5 & 90.6 & 89.9 & 92.3 & 91.9 & 87.3\\
        \hline
        \textbf{Ours} & \textbf{100} & \textbf{100} & \textbf{100} & \textbf{100} & \textbf{99.9} & \textbf{99.7} & \textbf{100} & \textbf{100} & \textbf{99.7} & \textbf{99.0} & \textbf{99.6} & \textbf{99.4} & \textbf{98.1} & \textbf{99.5} & \textbf{96.2} & \textbf{91.7}\\
        \bottomrule
    \end{tabular}
    \caption{Comparison of AUC scores (\%) of detecting gradient-based adv-faces from LFW and CelebA-HQ datasets. Detectors of LID and SID are trained on FGSM adv-faces. As for other detectors, all attacks are unseen. "RF." denotes RFGSM and "TIP." denotes TIPIM.}
    \label{tab:gradient-based}
\end{table*}

\subsection{Assumption Validation }
\begin{figure}[t]
    \centering
    \includegraphics[width=0.8\columnwidth]{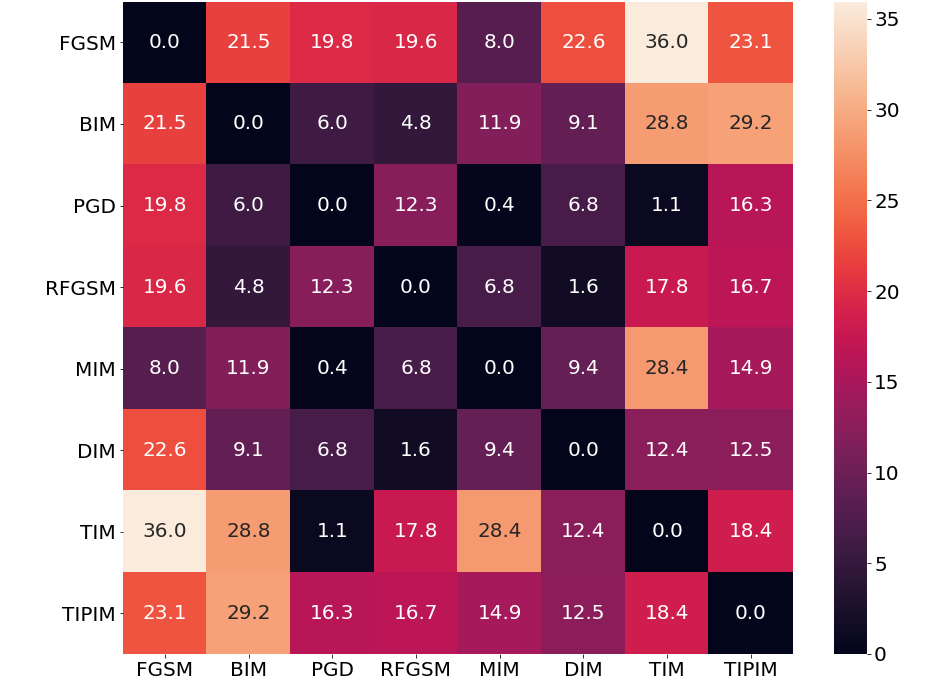}
    \caption{
    Fréchet Inception Distance heat map of various attacks.}
    \label{fig: FID}
\end{figure}

Our assumption for gradient-based noise patterns stands on the visual similarities between adversarial noises. However, the human eye is sometimes unreliable because we cannot observe tiny differences between pixels. 
To verify the reliability of the hypotheses, we train a simple XceptionNet with real faces from the CelebA-HQ dataset and adv-faces generated by one attack algorithm and test detection accuracy on 1,000 adv-faces per attack. As shown in Table \ref{tab: attack cross}, a detector for a specific gradient-based attack is able to generalize to other attacks except those based on GANs.
but the success rate on some attacks such as TIM is much lower. We also calculate the Fréchet Inception Distance of various attacks in Figure \ref{fig: FID} to visualize distribution similarity.
What the results demonstrate is that attack noises are to some extent similar to each other, and the way of extracting a universal noise pattern is feasible. 
The result also indicates that noise patterns of GAN-based attacks are not close to gradient-based attack noise patterns and need to be specifically treated. 

\subsection{Main Results}


\paragraph{Gradient-based adv-face detection}
We first compare the detection performance of our method to other detectors on gradient-based attacks. 
Baselines include 2 detectors proposed for adversarial example detection \cite{LID,SID} and 3 detectors for OOD detection \cite{ODIN,MD,ReAct}. 
The detectors are trained on real images, along with adv-faces images (LID and SID) or pseudo adv-faces images (Ours). 
We compute the classification AUC for all methods on a dataset comprising of 1k real images and 1k adversarial face images per attack type in LFW and CelebA-HQ datasets.

\begin{table}[ht]
    \centering
    \tabcolsep=0.3cm
    \begin{tabular}{l|ccc}
        \toprule
        Method & Adv-Makeup & AMT-GAN & Mean\\
        \hline
        He et al. & 52.5 & 88.2 & 70.4\\
        Luo et al. & 61.8 & 65.1 & 62.0 \\
        \hline
        ODIN & 63.3 & 69.7 & 66.5\\
        MD & 72.3 & 78.2 & 72.3 \\
        React & 77.6 & 82.9 & 80.3\\
        \hline
        \textbf{Ours} & \textbf{96.6} & \textbf{89.7} & \textbf{93.2}\\
        \bottomrule
    \end{tabular}
    \caption{Comparison of AUC (\%) of detecting GAN-based adv-faces from CelebA-HQ dataset. All attacks are unseen by all detectors. The detector of He et al. is pre-trained on CelebA-HQ and detector of Luo et al. is pre-trained on FF++ \protect\cite{FF++}. }
    \label{tab:GAN-based}
\end{table}

As shown in Table \ref{tab:gradient-based}, trained on the known attack (FGSM), previous adversarial example detection methods is difficult to detect unknown attacks. 
In a contrast, our method reaches a high level of detecting gradient-based adv-faces and almost approaches saturation performance on the LFW dataset.
This is likely because the network learns a generic representation for adv-faces. 
The result that the AUC score on LFW is higher than which on CelebA-HQ may contribute to the fact that CelebA-HQ is a more complicated dataset with high resolution and diversified background so self-perturbation works much harder on it.
Thus we speculate that a more complex and more diversified data environment will increase the difficulty of adv-face detection tasks.

\paragraph{GAN-based adv-face detection}
To investigate the advantage of our method on detecting GAN-based samples, we make comparisons with SOTA methods of face forgery detection \cite{YangHe2021,YuchenLuo2021} and OOD detection \cite{ODIN,MD,ReAct}. 
We train our model with only real faces and self-perturbed faces from CelebA-HQ and test both our approaches and baseline models on GAN-based adv-faces. 
As shown in Table \ref{tab:GAN-based}, the performance of our method exceeds which of other algorithms and models. 
This result verifies our assumption that GAN-made adv-faces have manipulated clues like an abnormal color aberration in high-frequency regions.
The observation that detection performance on Adv-Makeup is higher than which on AMT-GAN is likely because abnormal local aberrations of Adv-Makeup are more obvious than that of AMT-GAN.
Besides, the result shows that the detection of face forgery can not generalize to adv-faces although both deepfake faces and adv-faces are generated by GANs. 
This is possible because a huge difference exists between the fingerprints of deepfake GANs and that of adv-face GANs.


\subsection{Ablation}
\paragraph{Component ablation}
As argued, self-perturbation makes the detector learn a more generic representation of adv-faces, Max-Pooling Classifier captures abnormal local color aberrations, and the regularization term helps to boost adv-faces detection.
We then conduct an ablation study to verify the effectiveness of each component. 
As result shown in Table \ref{tab:ablation}, detectors trained without self-perturbation totally failed in detecting adv-faces from TIPIM and GAN-based attacks, indicating the indispensability of self-perturbation in detecting advanced adversarial attacks.
There is also an obvious gap between using and not using Max-Pooling Classifier, especially on GAN-based adv-faces.
Narrowing the decision boundary, the regularization term helps to filter out a small number of adversarial samples which are not similar enough to self-perturbed faces, further improving detection performance.
As for some gradient-based adv-faces such as DIM, detection accuracy is high enough regardless of using Max-Pooling Classifier and a regularization term.
It is likely because self-perturbation is similar enough to attack noise patterns so that the representations learned by networks are generic enough.

\begin{table}[t]
    \centering
    \tabcolsep=0.12cm
    \begin{tabular}{lccccccccc}
        \toprule
         Ablation & FGSM & DIM & TIM & TIP. & AdvM. & AMT.\\
        \hline
        w/o SP & 100 & 92.5 & 72.4 & 66.1 & 52.1 & 50.8 \\
        w/o MPC & 99.2 & 99.0 & \textbf{96.8} & 88.5 & 90.0 & 82.0 \\ 
        w/o LU & \textbf{99.7} & 99.0 & 95.5 & 91.2 & 95.1 & 88.4 \\ 
        \hline
        \textbf{SP+MPC+LU} & \textbf{99.7} & \textbf{99.5} & 96.2 & \textbf{91.7} & \textbf{96.6} & \textbf{89.7} \\ 
        \bottomrule
    \end{tabular}
    \caption{AUC (\%) comparison on CelebA-HQ dataset for ablation study. "w/o SP" means training on adv-faces generated by FGSM instead of self-perturbation. "MPC" refers to Max-Pooling Classifier. "LU" refers to training under the regularization $\mathcal{L}_{\text{uncertainty}}$. "AdvM." and "AMT." denote AdvMakeup and AMT-GAN respectively.}
    \label{tab:ablation}
\end{table}

\begin{table}[t]
    \centering
    \tabcolsep=0.15cm
    \begin{tabular}{l|ccccccc}
        \toprule
        Self-Perturbation & FGSM & TIM & TIP. & AdvM. & AMT.\\
        \hline
        Point-wise & 99.6 & 64.5 & 68.6 & 50.8 & 50.1\\
        Block-wise & 81.4 & 98.9 & 81.7 & 76.5 & 62.3\\
        GC & 65.5 & 70.7 & 68.0 & 85.5 & 78.0\\
        \bottomrule
    \end{tabular}
    \caption{Detection AUC (\%) on CelebA-HQ. Self-perturbed faces are generated using one of the self-perturbations. "GC" denotes self-perturbation for GAN-based attack mentioned in Algorithm \ref{alg:pseudo GAN}.}
    \label{tab: noise pattern cross}
\end{table}


\begin{table}[t]
    \centering
    \begin{tabular}{l|cccccccc}
        \toprule
        $\epsilon$ & 15 & 10 & 5 & 3 & 2 & 1\\
        \hline
        15 & 100 & 81.2 & 0.0 & 0.0 & 0.0 & 0.0 \\
        10 & 100 & 100 & 66.2 & 0.8 & 0.1 & 0.0 \\
        5 & 100 & 100 & 99.7 & 54.3 & 9.2 & 1.4 \\
        3 & 99.7 & 99.8 & 100 & 95.2 & 86.8 & 26.6 \\
        2 & 99.9 & 99.9 & 0.0 & 0.0 & 0.0 & 0.0 \\
        1 & 0.0 & 0.1 & 54.1 & 62.0 & 52.5 & 52.9 \\
        \bottomrule
    \end{tabular}
    \caption{Detection accuracy (\%) on adv-faces generated by FGSM with various of perturbation magnitudes. Networks are trained on adv-faces at $\epsilon/255$ in the left column, and tested on adv-faces produced at $\epsilon/255$ in the top row.}
    \label{tab: eps cross}
\end{table}

\begin{figure}[t]
	\centering
	\subfigure{
		\includegraphics[width=0.47\columnwidth]{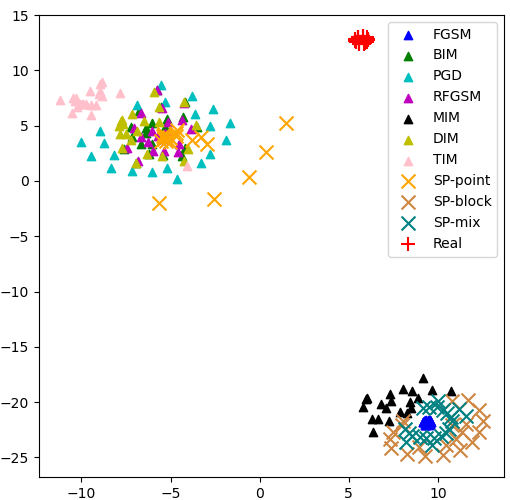} 
		\label{subfig: noise_tsne}}
	\subfigure{
		\includegraphics[width=0.47\columnwidth]{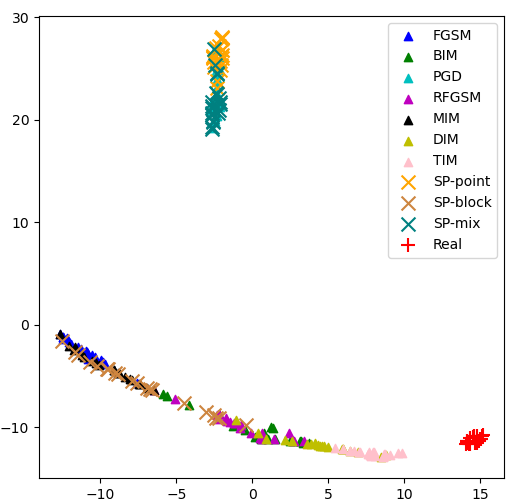}
		\label{subfig: real_adv_tsne}}
	\caption{2D t-SNE visualization. Left: flattened noise images of adv-faces and self-perturbed faces. 'SP-point' and 'SP-block' denotes point-wise and block-wise self-perturbations respectively, and 'SP-mix' means the mixture of the two perturbations. We use a zero-vector to represent the noise of real faces. Right: face features prior to the logit layer extracted by our model.}
	\label{fig: t-SNEs}
\end{figure}

\paragraph{Complementarity of self-perturbations}
To explore the necessity and complementarity of self-perturbations, we train a simple XceptionNet on self-perturbed faces generated from CelebA-HQ using only a single self-perturbation. 
The results are shown in Table \ref{tab: noise pattern cross}. As we can see, training a detector only relying on one of the self-perturbations is insufficient for detecting all unseen adv-faces. Due to the differences in generation procedures, self-perturbations are complementary to each other in the detection tasks. 

\subsection{Analysis of Our Approach}

\paragraph{Functionality of self-perturbations}

To explore the functionality of self-perturbations for gradient-based attacks, we extract noise features and face features of adversarial, self-perturbed, and real faces and visualize them using 2D t-SNE projection. 
As Figure \ref{fig: t-SNEs} shows, self-perturbation is very close to attack noise and far from zero-vector (real faces).
By separating real faces from self-perturbed faces, the trained network is able to distinguish between real and adv-faces indirectly. 
Although some adv-faces generated by TIM are not covered by self-perturbations, a model trained on decision boundary regularization still separates them from real faces.



\paragraph{Impact of hyper-parameter}
Another experiment is about the choice of $\epsilon$. We train a simple XceptionNet on FGSM adv-faces at $\epsilon = 1/255$ to $\epsilon = 15/255$ and test on adv-faces at different $\epsilon$. As shown in Table \ref{tab: eps cross}, detectors trained on a smaller $\epsilon$ are able to detect adv-faces generated with a larger $\epsilon$. But an extremely small $\epsilon$ may cause failure in model training. In practice, we do not need to select an extremely small value for $\epsilon$ because the attack success rate is too low to take into account as shown in the Appendix. In consideration of this, we choose $\epsilon = 5/255$ (attack success rate lower than 1\%) for producing and detecting adv-faces. 

\section{Conclusion}
In this paper, we investigate the intrinsic similarities of recent adv-faces and heuristically design three kinds of real face self-perturbations close to attack noise pattern. 
Regarding adv-faces as OOD data, we propose an FRS-agnostic and attack-agnostic cascaded system for adv-faces detection, which includes real face self-perturbations, decision boundary regularization, and Max-Pooling Classifier focusing on abnormal local color aberrations.
Trained on only real faces and self-perturbed real faces, our model learned generic representations for adv-faces.
Comprehensive analysis validates the proposed assumptions that noises of adv-faces have intrinsic similarities and exist in high-frequency areas, and extensive experiments demonstrate the improved effectiveness of our method compared with several recent baselines.
Although our method achieves a satisfactory performance on detecting adv-faces (almost approaches saturation on LFW dataset), we observe that the latent representation learned by the detector is hard to generalize to other domains. For instance, the detector trained on the LFW dataset fails to detect adv-faces generated from the CelebA-HQ dataset. 
For future work, we will try to solve this matter, by learning a more generic representation across various domains.

\section{Acknowledgement}

This work was supported in part by the Natural Science Foundation of China under Grant 61972169, in part by the National key research and development program of China(2019QY(Y)0202, 2022YFB2601802), in part by the Major Scientific and Technological Project of Hubei Province (2022BAA046, 2022BAA042), in part by the Research Programme on Applied Fundamentals and Frontier Technologies of Wuhan(2020010601012182) and the Knowledge Innovation Program of Wuhan-Basic Research, in part by China Postdoctoral Science Foundation 2022M711251.

\bibliographystyle{named}
\bibliography{ijcai23_wq}

\begin{appendices}
\section{Experimental details}
In this section, we specify the procedure of real face self-perturbations including hyper-parameters, pipelines, and experiment equipment. 
\subsection{Implementation Details of Self-Perturbation}
Real face self-perturbations are produced by using some random parameters. 
As for self-perturbations for the gradient-based attack, we use random noise value $\alpha \in [-5/255, 5/255]$ with perturbation magnitude $\epsilon = 5/255$. 

\begin{figure*}[t]
    \centering
    \includegraphics[width=2\columnwidth]{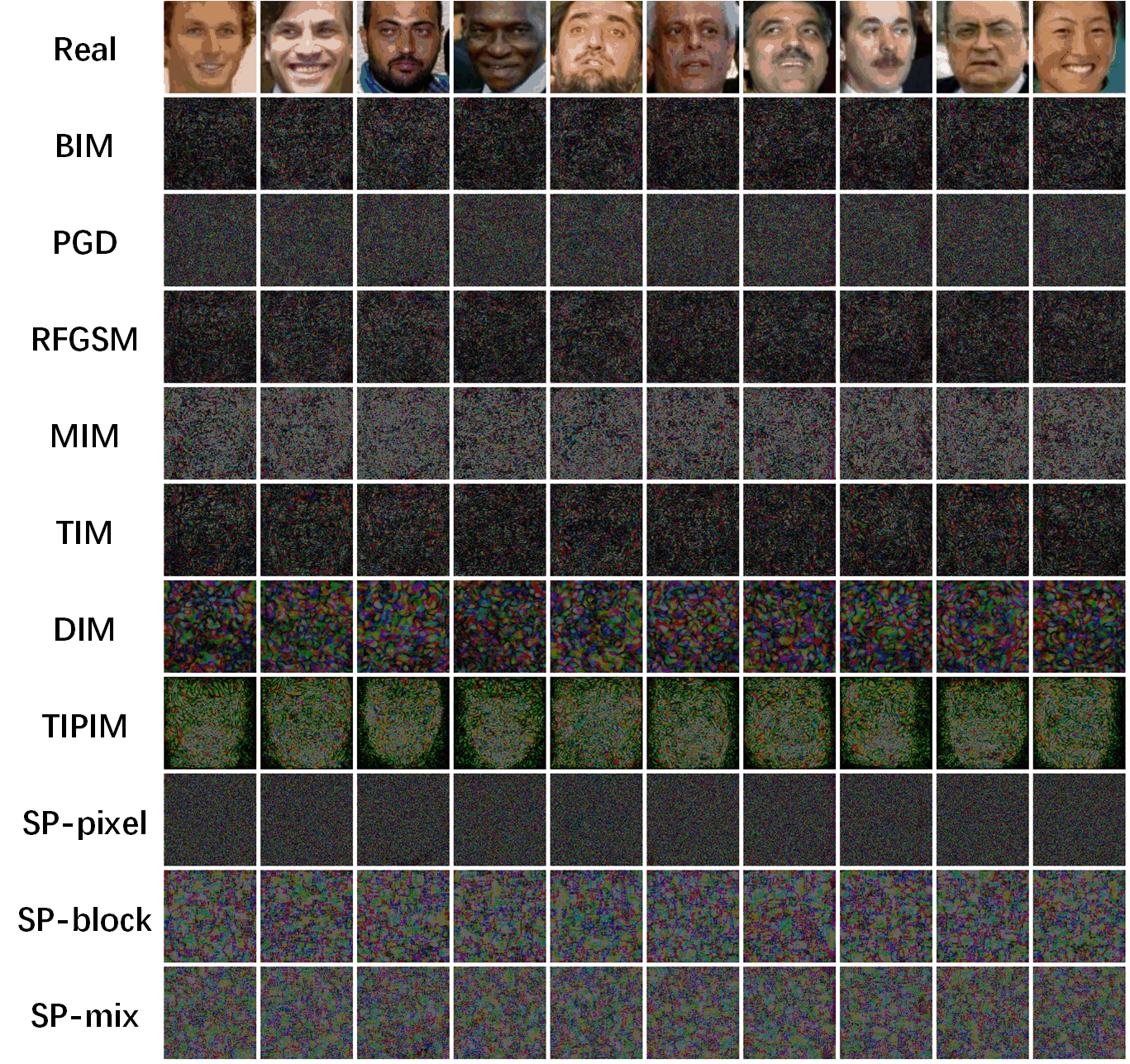}
    \caption{Noise images of gradient-based attacks and self-perturbations. All images are generated at $\epsilon=10$. Values of noise images are 10 times expanded for better visualization.}
    \label{fig: Res}
\end{figure*}

For each self-perturbation for the gradient-based attack, $\textrm{mode}$ is randomly selected where $\textrm{mode} \in \{\textrm{"point"},\textrm{"block"},\textrm{"mix"}\}$ and "mix" denotes the summation of "pixel" and "block" perturbations before clip operation. Noise patterns of gradient-based attack and self-perturbation are shown in Figure \ref{fig: Res}.

\begin{figure}[t]
    \centering
    \includegraphics[width=0.95\columnwidth]{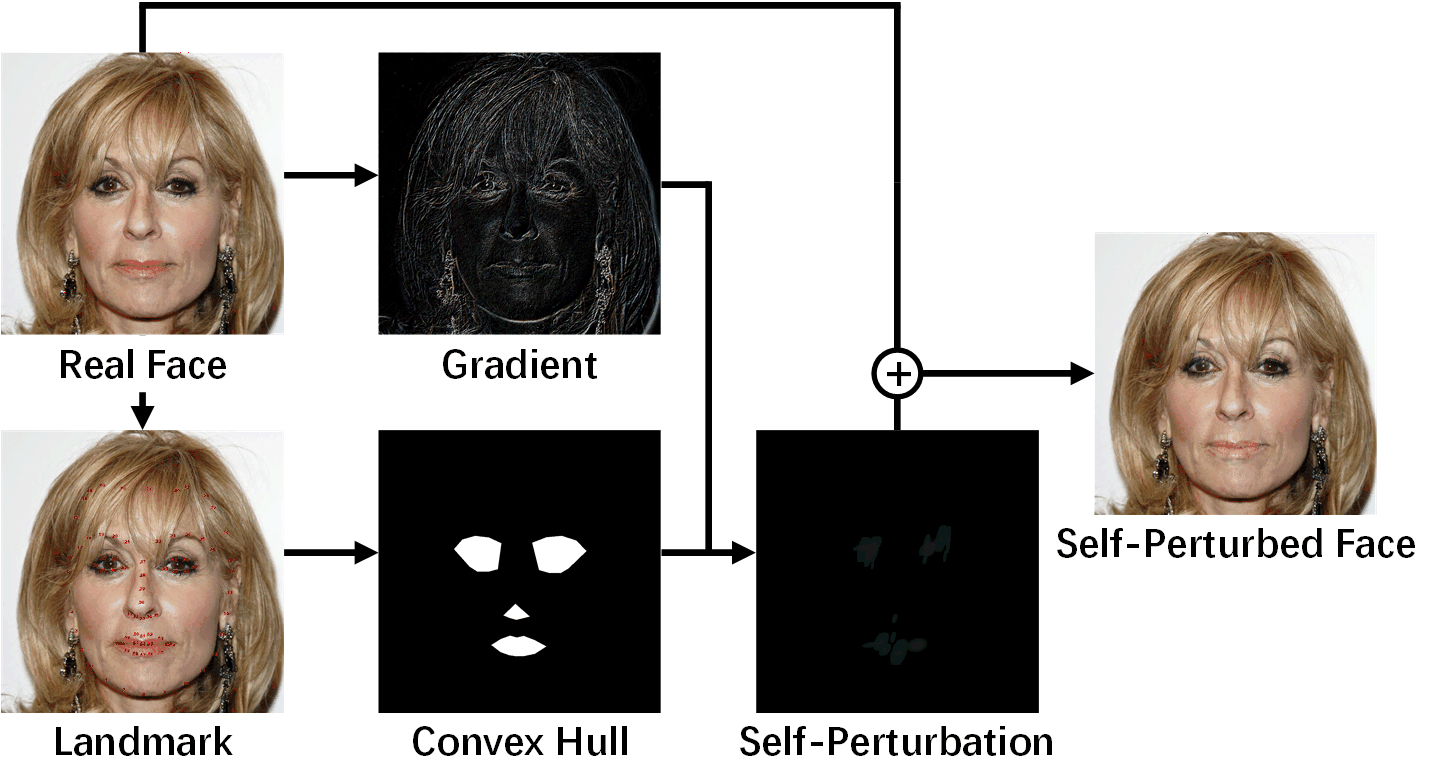}
    \caption{Pipeline of generating self-perturbations for GAN-based attack.}
    \label{fig: SP_GC}
\end{figure}

\begin{figure*}[t]
    \centering
    \includegraphics[width=2\columnwidth]{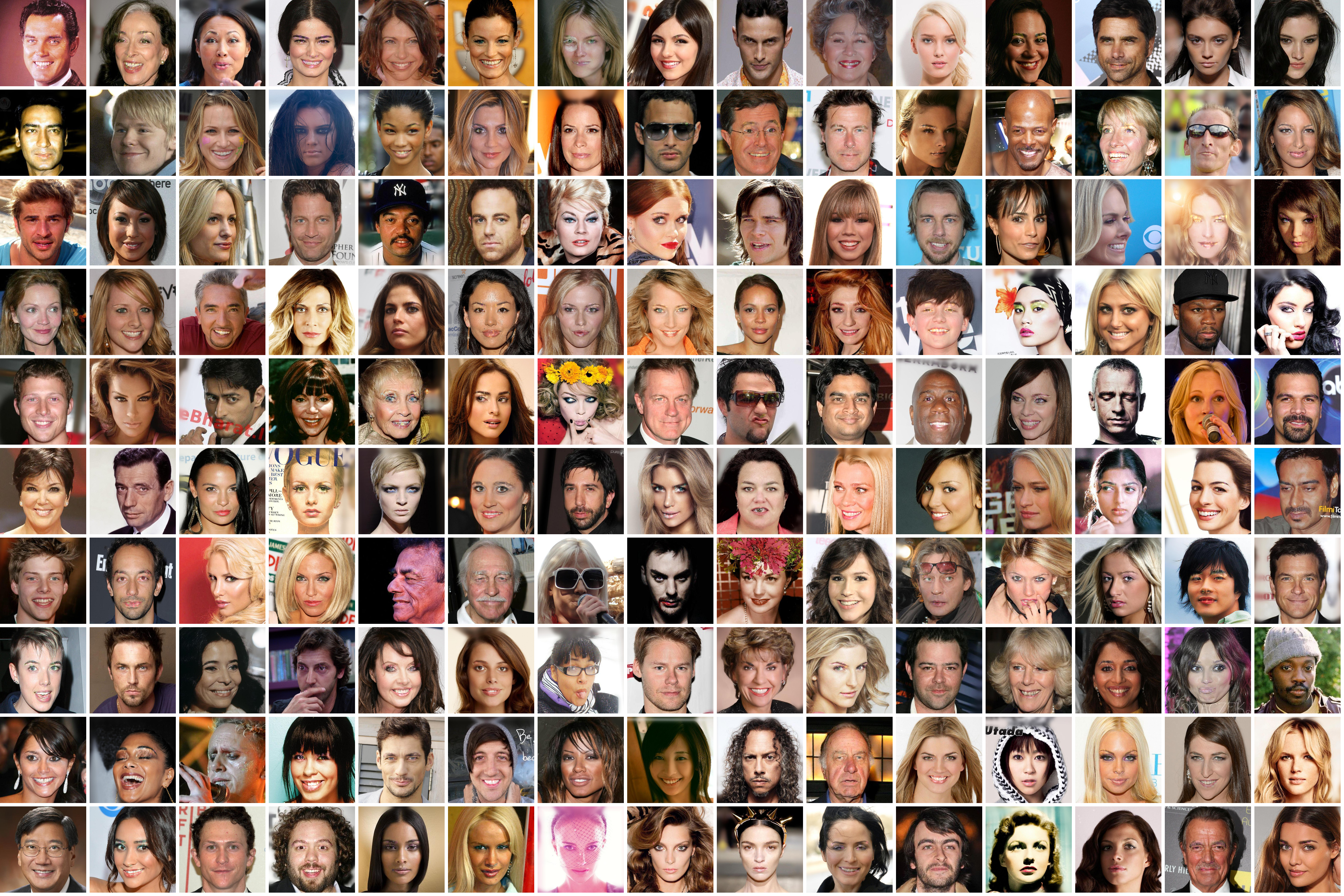}
    \caption{Self-perturbed faces for GAN-based attacks. All real face images are sampled from the CelebA-HQ dataset.}
    \label{fig: SP_GCselect}
\end{figure*}

As for self-perturbations for GAN-based attacks, we set random perturbation magnitude to $\epsilon \in [10,70]$. 
Real face images are sampled from the CelebA-HQ dataset with a size of $1024\times 1024$ and then resized to $256\times 256$. 
A pixel from a convex hull is added to the high-frequency set $H$ if its value is over a threshold of 50, and selected to the subset $H_s$ by a probability in $[0.016, 0.040]$. 
Each gradient color (GC) patch is generated in a random size in $[2,25]\times[2,25]$.
Before adding GC patches to high-frequency pixels, we randomly select a GC patch and add it to each pixel by a random probability of $0.8$, and add other GC patches to the rest of the pixels. The high-frequency areas are the eyes, nose tip, and mouse. 
We portray the procedure of generating self-perturbations for GAN-based attack in Figure \ref{fig: SP_GC}, and exhibit a set of them in Figure \ref{fig: SP_GCselect}.

\subsection{Software and Hardware}
We run all experiments with Python 3.8.8 and PyTorch 1.10.0, using two NVIDIA GeForce RTX 1080 GPUs. 

\begin{table}[b]
    \centering
    \tabcolsep=0.07cm
    \begin{tabular}{c|c|ccccccc}
        \toprule
        $\epsilon$ & FAR & FGSM & BIM & PGD & RFGSM & MIM & DIM & TIM \\
        \midrule
        \multirow{3}{*}{ }5 & 0.01 & 0.03 & 0.05 & 0.11 & 0.15 & 0.41 & 0.84 & 0.09\\
        \multirow{3}{*}{ } & 0.001 & 0.01 & 0.08 & 0.24 & 0.43 & 0.90 & 1.72 & 0.15\\
        \hline
        \multirow{3}{*}{ }10 & 0.01 & 0.05 & 29.52 & 25.61 & 23.21 & 42.28 & 82.83 & 25.73\\
        \multirow{3}{*}{ } & 0.001 & 0.08 & 38.05 & 34.01 & 31.63 & 50.25 & 87.44 & 34.09\\
        \hline
        \multirow{3}{*}{ }15 & 0.01 & 0.25 & 44.00 & 35.26 & 59.07 & 70.22 & 96.85 & 65.11\\
        \multirow{3}{*}{ } & 0.001 & 0.83 & 50.44 & 42.35 & 67.48 & 76.96 & 98.32 & 73.08\\
        \hline
        \multirow{3}{*}{ }20 & 0.01 & 2.15 & 76.86 & 69.08 & 74.20 & 83.24 & 99.47 & 83.31\\
        \multirow{3}{*}{ } & 0.001 & 5.17 & 83.03 & 76.76 & 81.81 & 88.03 & 99.80 & 88.54\\
        \bottomrule
    \end{tabular}
    \caption{Attack success rate @FAR (\%) of various of gradient-based adversarial attack methods at different $\epsilon/255$.}
    \label{tab: ASR}
\end{table}

\section{Attack Success Rate}
To better explain the reason why we use $\epsilon = 5/255$ in our experiments, we record the attack success rate (ASR) of FGSM, BIM, PGD, RFGSM, MIM, DIM, and TIM at different perturbation magnitude $\epsilon$. Perturbation magnitude is set to $\epsilon \in \{5,10,15,20\}$, the learning rate is set to $\alpha = 5$, and the amount of iteration steps is set to $10$. The victim facial recognition system (FRS) is ArcFace. False Accept Rate (FAR) for threshold selection is set to 0.01 and 0.001, and thresholds are computed as 27.25\% and 31.99\% respectively. Results are conducted on 7,484 adversarial faces (adv-faces) produced by attacking ArcFace. 
As shown in Tabel \ref{tab: ASR}, ASR under $\epsilon=5/255$ is a very small value to take into account, so we choose $\epsilon = 5/255$ for adv-faces detection evaluation. 

All above ASR is calculated at $\alpha=5$ and $\textrm{steps}=5$. However, a smaller learning rate and larger steps may slightly promote the attack performance because a smaller learning rate will make attacks more sufficient especially when the perturbation is subtle. In view of this, all of adv-faces used in our main results are produced at $\alpha=1$ and $\textrm{steps}=10$. 

\section{Validation of Intrinsic Similarities}

\begin{figure}[b]
    \centering
    \includegraphics[width=0.8\columnwidth]{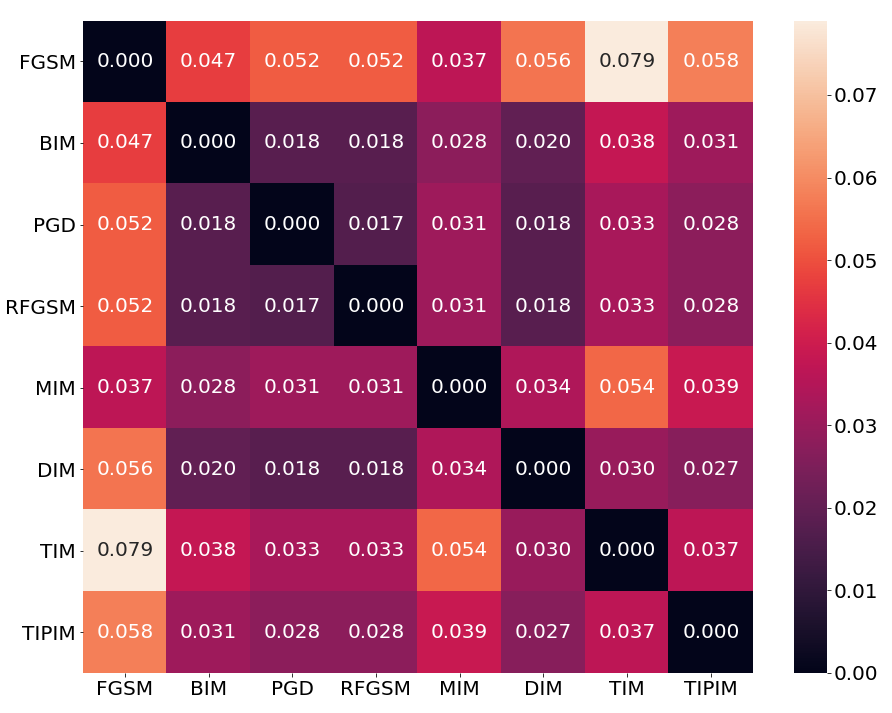}
    \caption{
    Learned Perceptual Image Patch Similarity (LPIPS) heat map of various attacks.}
    \label{fig: LPIPS}
\end{figure}

Besides the FID metric reported in the main paper, we test the Learned Perceptual Image Patch Similarity (LPIPS) and draw a heat map to illustrate the similarities between various attacks. The heat map is shown in Figure \ref{fig: LPIPS}.

\begin{figure}[t]
    \centering
    \includegraphics[width=0.8\columnwidth]{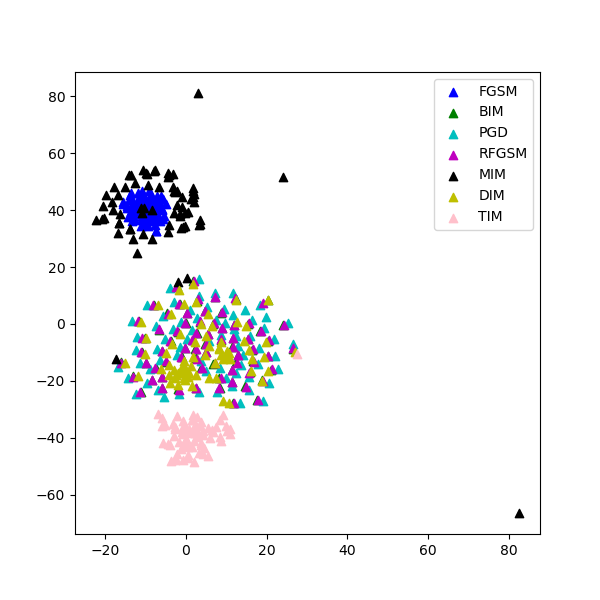}
    \caption{
    2D t-SNE visualization of K-Means clusters.}
    \label{fig: tsne_adv}
\end{figure}

Another quantitative method is about K-Means. We conduct a K-Means experiment on 7 gradient-based attacks at $K=7$.
If there is no similarity exists between these attacks, each attack would form a cluster and there is no intersection between the clusters. Otherwise, similarity exists.
As shown in Table \ref{tab: kmeans}, noises from one attack always occur in two or more clusters, such as BIM contributes 26.9\% to cluster 2, 45.5\% to cluster 4, and 37.7\% to cluster 5; RFGSM contributes 28.0\% to cluster 2, 45.5\% to cluster 4, and 36.8\% to cluster 5. 
Clusters are visualized in Figure \ref{fig: tsne_adv}.
It demonstrates the existing latent similarities that clusters always include noises from more than one attack.

\begin{table}[t]
    \centering
    \tabcolsep=0.09cm
    \begin{tabular}{c|cccccccc}
        \toprule
        Cluster & FGSM & BIM & PGD & RFGSM & MIM & DIM & TIM \\
        \midrule
        0 &  &  & 98.6 &  &  &  & 1.4 \\
        1 & 100 &  &  &  &  &  &  \\
        2 &  & 26.9 &  & 28.0 &  & 45.1 &  \\
        3 &  &  &  &  &  &  & 100 \\
        4 &  & 45.5 &  & 45.5 &  & 9.0 &  \\
        5 &  & 37.7 &  & 36.8 &  & 25.5 &  \\
        6 &  &  &  &  & 100 &  &  \\
        \bottomrule
    \end{tabular}
    \caption{K-Means experimental result of adversarial attacks at $K=7$. Each value indicates the proportion of an attack in a cluster.}
    \label{tab: kmeans}
\end{table}

\begin{table}[h]
    \centering
    \definecolor{Gray}{gray}{0.9}
    \tabcolsep=0.05cm
    \begin{tabular}{c|cc|c|cc}
        \toprule
        Classifier & AdvM. & AMT. & OOD & AdvM. & AMT.  \\
        \midrule
        Avg-Pooling & 89.8 & 81.9 & GODIN & 72.0 & 79.5  \\
        Voting & 87.5 & 80.4 & LogitNorm & 69.4 & 63.4  \\
        \rowcolor{Gray} Max-Pooling & 95.1 & 88.4 & LU & 90.0 & 82.0  \\
        \bottomrule
    \end{tabular}
    \caption{Ablation on classification strategies and OOD frameworks. 
    The values indicate the detection AUC($\%$) using XceptionNet, the higher the better. 
    The last row represents our optimal configuration. 
    ”LU” denotes our Decision Boundary Regularization.
    "AdvM." and "AMT." denote AdvMakeup and AMT-GAN respectively.
    }
    \label{tab: detail ablation}
\end{table}

\section{Detailed Ablation}
In this section, we conduct ablation studies of classification mechanisms and regularization methods.  
The corresponding experiments of detecting GAN-made adv-faces on the CelebA-HQ dataset are shown in Table \ref{tab: detail  ablation}, which supports the improved effectiveness of our method. The analyses are as follows.
\subsection{Classification Mechanism}

Our Max-Pooling Classifier is designed to focus on local abnormal color variations which appear in GAN-made adv-faces. 
Once an anomaly appears somewhere, the entire image is considered an adv-face.
We conduct an ablation study of the mechanism of the classifier compared with the average-pooling mechanism and voting mechanism.
As shown in Table \ref{tab: detail ablation}, max-pooling outperforms others because the average strategy would dilute the impact of local abnormal color variations, and a voting mechanism would backfire in case most regions are normal.

\subsection{Regularization}
Our Decision Boundary Regularization aims to shrink the decision boundary towards real faces so that adv-faces closed to the real are more likely to be detected.
We compared our methods with training strategies of GODIN\cite{GODIN} and LogitNorm\cite{logitnorm}, and it turns out more effective than other existing OOD frameworks.

\section{Facial Edition Attack and Natural Dataset}

In this section, we test our methods against facial attribute attacks which are likely used in the physical world.
We also test our methods by detecting adversarial examples from the general dataset to validate its independence from datasets.
\subsection{Facial Attribute Attack Detection}
We test our model by detecting three facial attribute editing attacks: sticker attack\cite{advmask}, eyeglass attack\cite{advglass}, and face mask attack\cite{advmask} on the VGGFace2 dataset.
The results shown in Table \ref{tab: facial} demonstrate the validity of our model.

Compared to GAN-based attacks, evidence of facial attribute editing attacks are easier to be detected since the editing traces are more obvious. 
Our hypotheses still hold for them because they also perturb high-frequency regions.

\subsection{Experiment on General Dataset}
We conduct experiments on CIFAR10 and ImageNet datasets, using only self-perturbation for the gradient-based attack.
CIFAR10 is a general dataset with low resolution, while the resolution of ImageNet is higher.
Consistent with the main paper, perturbation magnitude is set to $\epsilon = 5/255$ for both CIFAR10 and ImageNet.
It is smaller than $\epsilon = 12.75/255$ used in baseline LID \cite{LID}for CIFAR10 and $\epsilon = 16/255$ used in LiBRe \cite{LiBRe} for ImageNet, which means our setting is more challenging.

As shown in Table \ref{tab: general}, our self-perturbation still works on general datasets with both lower and higher resolution.
It shows that our method is domain-agnostic.

\begin{table}[hb]
    \centering
    \definecolor{Gray}{gray}{0.9}
    \begin{tabular}{c|ccc}
        \toprule
        Method & Sticker$[3]$ & Eyeglass$[4]$ & Face Mask$[3]$  \\
        \midrule
        LID & 51.0 & 64.7 & 53.2  \\
        SID & 61.5 & 59.1 & 57.5   \\
        \rowcolor{Gray}Ours & 99.8 & 92.4 & 98.6   \\
        \bottomrule
    \end{tabular}
    \caption{AUC scores ($\%$) of detecting facial editing attacks.}
    \label{tab: facial}
\end{table}

\begin{table}[hb]
    \centering
    \tabcolsep=0.09cm
    \begin{tabular}{c|ccccccc}
        \toprule
        Dataset & FGSM & BIM & PGD & RF. & MIM & DIM & TIM  \\
        \midrule
        CIFAR10 & 99.8 & 99.7 & 99.8 & 99.8 & 99.6 & 99.8 & 94.4  \\
        ImageNet & 99.3 & 98.6 & 98.8 & 98.6 & 98.3 & 99.3 & 98.3  \\
        \bottomrule
    \end{tabular}
    \caption{Detection AUC ($\%$) on general image datasets.}
    \label{tab: general}
\end{table}

\end{appendices}

\end{document}